%% file: ijcv.tex
\RequirePackage{fix-cm}
\documentclass[twocolumn]{svjour3}          
\smartqed  
\usepackage{graphicx}
\usepackage{epsfig}
\usepackage{amsmath}
\usepackage{amssymb}

\usepackage{soul}
\usepackage[utf8]{inputenc}
\usepackage[pagebackref=true,breaklinks=true,letterpaper=true,colorlinks,bookmarks=false,citecolor=cyan]{hyperref}

\usepackage[figuresright]{rotating}
\usepackage{textcomp,subfigure,multirow,upgreek,tabularx}
\usepackage{graphics,gensymb}
\usepackage{bm,cases,threeparttable}

\usepackage{setspace}

\graphicspath{{Figures/}}
\usepackage{booktabs}

\usepackage{amsfonts}
\def\R{\mathbb{R}} 
\usepackage{caption}
\usepackage[table]{xcolor}
\usepackage{dirtytalk}
\usepackage{appendix}
\usepackage{stfloats}

\usepackage{algorithm}
\usepackage{algpseudocode}

\newcommand{\algrule}[1][.2pt]{\par\vskip.2\baselineskip\hrule height #1\par\vskip.2\baselineskip}

\usepackage{enumitem}
\setitemize{noitemsep,topsep=5pt,parsep=0pt,partopsep=0pt}

\newenvironment{itemizetight}
{ \begin{itemize}
    \setlength{\itemsep}{0pt}
    \setlength{\parskip}{0pt}
    \setlength{\parsep}{0pt}     
    \setlength{\topsep}{0pt}      
    \setlength{\partopsep}{0pt}      }
{ \end{itemize}                  }

\begin{document}

\title{Pre-training for Action Recognition with Automatically Generated Fractal Datasets
}


\author{Davyd Svyezhentsev \and
        George Retsinas \and
        Petros Maragos 
}


\institute{
	Davyd Svyezhentsev (dsvyez@gmail.com) \\
	George Retsinas (gretsinas@central.ntua.gr) \\
	Petros Maragos (maragos@cs.ntua.gr) \\\\
	School of Electrical and Computer Engineering, National Technical University of Athens, Athens, Greece \\	
}

\date{Received: date / Accepted: date}

\maketitle

\input{0-abstract}
\input{1-introduction}
\input{2-proposed}

\input{3-experiments}
\input{4-related-work}
\input{5-conclusions}

\input{ref}

\input{6-appendix}

%
%



\end{document}

%% file: 0-abstract.tex
\begin{abstract}

In recent years, interest in synthetic data has grown, particularly in the context of pre-training the image modality to support a range of computer vision tasks, including object classification, medical imaging etc.
Previous work has demonstrated that synthetic samples, automatically produced by various generative processes, can replace real counterparts and yield strong visual representations. This approach resolves issues associated with real data such as  collection and labeling costs, copyright and privacy.
    
We extend this trend to the video domain applying it to the task of action recognition. Employing fractal geometry, we present methods to automatically produce large-scale datasets of short synthetic video clips, which can be utilized for pre-training neural models. 
The generated video clips are characterized by notable variety, stemmed by the innate ability of fractals to generate complex multi-scale structures.
To narrow the domain gap,  we further identify key properties of real videos and carefully emulate them during pre-training. 
Through thorough ablations, we determine the attributes that strengthen downstream results and offer general guidelines for pre-training with synthetic videos. 
The proposed approach is evaluated by fine-tuning pre-trained models on established action recognition datasets HMDB51 and UCF101 as well as four other video benchmarks related to group action recognition, fine-grained action recognition and dynamic scenes. Compared to standard Kinetics pre-training, our reported results come close and are even superior on a portion of downstream datasets. \input{data_availability_statement}

\end{abstract}

%% file: data_availability_statement.tex
Code and samples of synthetic videos are available at \url{https://github.com/davidsvy/fractal_video}.

%% file: 1-introduction.tex
\section{Introduction}

Contemporary computer vision models require enormous amounts of data for training. Beginning with  ImageNet  \cite{russakovsky1} which consists of 1.4 million labeled images, the scale of vision datasets has been rapidly increasing, reaching tens of millions to a billion of samples \cite{goyal1,yalniz1,ghadiyaram1}. Regarding such datasets, multiple issues emerge. First,  data collection and annotation is arduous and expensive. Second, it has been noted that vision datasets may inherit human biases \cite{steed1,buolamwini1,wilson1,zhao1} and contain inappropriate content \cite{prabhu1}. Third, the depiction of humans in these datasets poses questions of privacy \cite{asano1}. Lastly, ownership concerns limit many datasets to noncommercial usage only.

Hence, the computer vision society has recently exhibited growing interest in synthetic datasets that mitigate the aforementioned shortcomings. Amongst them, noteworthy is the seminal work of \cite{kataoka1} who proposed to pre-train 2D CNNs with automatically generated images of fractals \cite{barnsley1}. Although their results are inferior compared to standard ImageNet pre-training, they significantly surpass training from scratch. Subsequent work has either enhanced their approach \cite{baradad1,anderson1,kataoka2}, alleviating the gap in downstream results between real and synthetic data or extended it to other domains \cite{yamada1}. Such datasets offer several advantages. They are constructed automatically and thus, do not require a collection or an annotation stage. They are also not bound by copyright limitations. Moreover, there are no questions regarding biases, inappropriate content or privacy, as no human subjects are depicted.

The present work extends the ideas of \cite{kataoka1} to the domain of video. Specifically, we seek to automatically produce synthetic datasets that are suitable for pre-training neural networks for the task of action recognition. Automatic action recognition is of paramount importance as it enables accurate detection and interpretation of human actions from video or sensor data. This technology has broad applications across various sectors, including surveillance, healthcare, robotics, sports analysis, and human-computer interaction. The significance of action recognition is additionally outlined by the sheer amount of videos available on the internet. With over 500 hours of video uploaded to YouTube every minute, there is an immediate need for robust algorithms that can help organize, summarize and retrieve this massive amount of data.

\begin{figure*}[!t]
    \includegraphics[width=\linewidth]{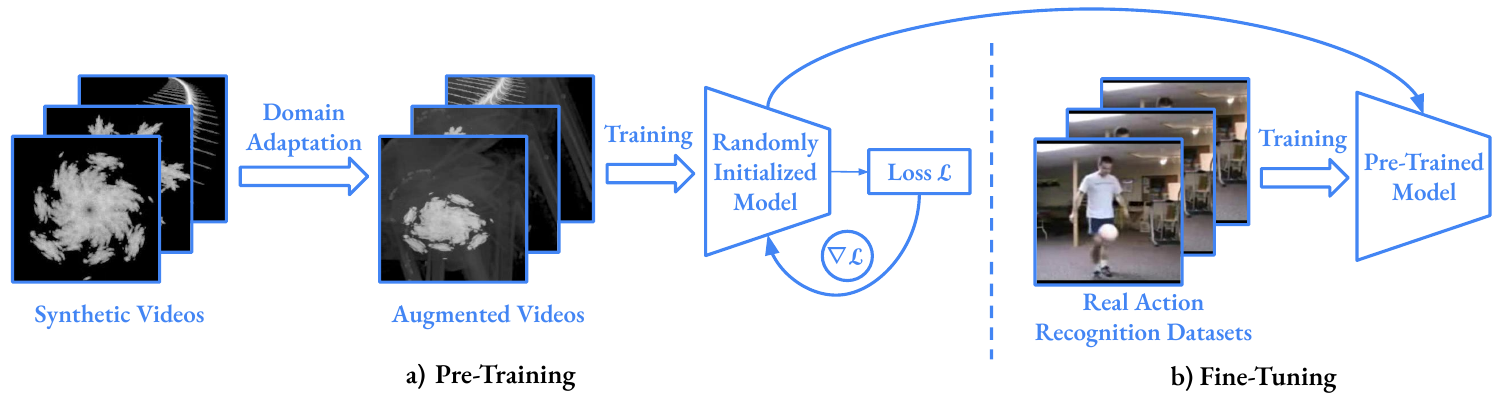}
    \caption{Overview of the proposed approach. Aiming to pre-train neural models, we utilize fractal geometry and automatically construct large-scale datasets of short synthetic video clips (Sec. \ref{sub:fractal_animation}). We additionally narrow the domain gap between real and synthetic videos by identifying key properties of the former and emulating them during pre-training (Sec. \ref{sub:domain_adaptation}). The transferability of the proposed datasets and transformations is experimentally assessed by fine-tuning the pre-trained models on real action recognition benchmarks (Sec. \ref{sec:experiments}) 
    }
    \label{fig:overview}
\end{figure*}

 Our approach is summarized in Fig. \ref{fig:overview}, where the two involved stages are depicted: 1) the \emph{pre-training} stage and 2) the \emph{fine-tuning} stage.
 The former includes the generation of fractal-based videos, followed by their appropriate augmentation to simulate human actions, as well as the training procedure to capture as much relevant information as possible.
 The latter includes a straightforward fine-tuning pipeline in order to adapt the pre-trained network to real action recognition datasets.   

The main contributions of this paper are:

\begin{itemizetight}
    \item Using fractal geometry \cite{barnsley1}, as well as other generative processes, we propose a pipeline that can automatically construct large-scale datasets of short synthetic video clips. These datasets are employed for pre-training neural networks for the task of action recognition instead of the typical large-scale Kinetics \cite{carreira2,kay1} dataset. 
    Both supervised and self-supervised learning is applicable 
    and explored in the experimental section.

    \item 
    Starting from the observation of real video samples, we pinpointed their fundamental attributes such a periodic motion, random background, camera displacement etc. These attributes are carefully emulated during pre-training, significantly reducing the domain gap between synthetic and real videos.

    \item Experimentally, we analyze downstream performance as a function of the training objective and the properties of the synthetic dataset. We determine beneficial attributes and offer general guidelines for pre-training with synthetic videos.

    \item We conduct error analysis of the pre-trained models' predictions and detect common patterns. As such, we propose tailored modifications to the synthetic data that may further boost downstream results in future work.

\end{itemizetight}

%% file: 2-proposed.tex
\section{Proposed Methodology}

\subsection{Preliminaries: Fractal Images}
\label{sub:fractal_intro}

Before exploring the video modality, it is necessary to first examine the simpler domain of images. Following \cite{kataoka1}, who produce synthetic image datasets to pre-train 2D CNNs, fractals generated via the Iterated Function Systems (IFS) technique \cite{barnsley1} are chosen as the backbone of our work.  
Specifically, the IFS fractals possess a set of appealing properties, including an easily implementable rendering algorithm as well as the possibility of producing a near-limitless supply of diverse images by randomly sampling parameters.

\begin{figure*}[!t]
    \includegraphics[width=\linewidth]{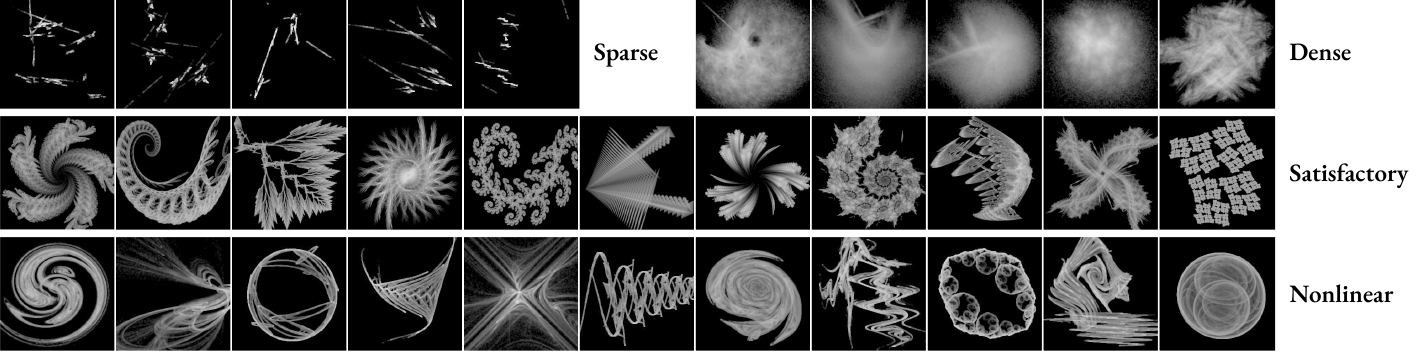}
    \caption{Examples of rendered 2D IFS attractors. A subset of linear samples exhibits unsatisfactory geometry, being either too sparse or too dense. Adding nonlinearity significantly alters the distribution of produced images and boosts overall diversity.}
    \label{fig:attractor}
\end{figure*}

\subsubsection{Classic IFS}
\label{subsub:classic_ifs}

Following Barnsley \cite{barnsley1}, a 2D IFS can be defined as a set of $N > 1$ affine transformations $F_i: \mathbb{R} ^ 2 \mapsto \mathbb{R} ^ 2$. Each $F_i$ is represented by a matrix $A_i \in \mathbb{R}^{2 \times 2}$ and a vector $b_i \in \mathbb{R} ^ 2$: 

\begin{align*}
F_{i}(\boldsymbol{x}; A_i, b_i) = A_i \boldsymbol{x} + b_i = 
 \begin{bmatrix}
 a_{i} & b_{i} \\
 d_{i} & e_{i} \\
 \end{bmatrix}
 \boldsymbol{x} +
 \begin{bmatrix}
 c_{i} \\
 f_{i} \\
 \end{bmatrix}
\end{align*}

The complete parameter matrix will be referred to as $W \in \mathbb{R}^{N \times 6}$. Additionally, it is necessary for each mapping to be  contractive with respect to the Euclidean distance $\mathrm{d}(\makebox[1ex]{\textbf{$\cdot$}} , \makebox[1ex]{\textbf{$\cdot$}})$:

\begin{align*}
d(F_i(x), F_i(y) ) \leq k_i \cdot \mathrm{d}(x, y), 0 < k_i < 1
\end{align*}

This constraint prevents divergence and ensures that, after successive applications of the mapping, points get progressively closer together. Furthermore, if the affine transformations are extended to be applied to whole subsets of the plane instead of single points, then their union $F$ is a contractive mapping with respect to the Hausdorff distance  on the space of nonempty compact sets \cite{barnsley1}. If $F$ is iteratively applied, starting from an arbitrary initial set, the iterations will converge to a unique fixed set of points which is referred to as the attractor of the IFS \cite{hutchinson1}. Since we are working with the 2D Euclidean plane, the resultant attractor is an image.

An approximation of the attractor (Fig. \ref{fig:attractor}) can be rendered with the chaos game algorithm \cite{barnsley1}. At first, this algorithm initializes the output image as zeros and samples a starting 2D point. At each iteration, one of the $N$ functions is sampled from the IFS and applied to the said point. The probability of selecting each function is: 

\begin{align*}
p_i = \frac{|\mathrm{det}(A_i)|}{\sum_{j=1}^{N} |\mathrm{det}(A_j)|}
\end{align*}

Given that the coordinates of the point are real numbers, they are quantized to a discrete pixel. 
The output image value corresponding to this this pixel is then incremented by one. 
After completing a specified number of iterations, the output, which is a 2D histogram, is normalized, producing a grayscale image.

To construct a dataset of fractal images, \cite{kataoka1} originally proposed to sample parameters independently from $U(-1, 1)$. However, a subsequent work \cite{anderson1} observed that this strategy often results in images with degenerate geometry, being either too sparse or too dense. As a solution, it is suggested to decompose each weight matrix into $A = R_{\theta} \Sigma R_{\phi} D$, omitting the index $i$ for brevity. The decomposed matrices are: 

\begin{itemizetight}
    \item $R_{x}$ is a rotation matrix parameterized by angle $x$.

    \item $\Sigma$ is a diagonal matrix containing the singular values $\sigma_1$ and $\sigma_2$ ordered by decreasing magnitude.

    \item $D$ is a diagonal matrix with elements $d_1, d_2 \in \{-1, 1\}$, acting as a reflection matrix.
\end{itemizetight}

As such, a ``well-behaved" $A$ can be sampled by appropriately sampling the decomposed parameters $\{\theta, \phi \allowbreak,  \allowbreak \sigma_1, \allowbreak \sigma_2, \allowbreak d_1, d_2\}$. Notably, \cite{anderson1} empirically deduce that the geometry of the resultant attractor depends on the quantity $ a = \sum_{i=1}^{N} (\sigma_{i, 1} + 2 \sigma_{i, 2})$, with  unsatisfactory behavior being minimized when:

\begin{align*}
    a_l = \frac{1}{2} (5 + N) \leq a \leq a_u = \frac{1}{2} (6 + N)
\end{align*}

Towards satisfying the inequality, the authors of \cite{anderson1} also propose an iterative algorithm for sampling parameters that fulfill the specified inequality.
Their methodology is adopted throughout the present work.

\subsubsection{Fractal Flame}
\label{sub:fractal_flame}

The Fractal Flame Algorithm \cite{draves1} is an extension to the ordinary IFS designed  to generate more aesthetically pleasing images, which are often employed as desktop backgrounds. Although multiple modifications to the IFS are introduced, only one is of interest for this work: nonlinearity. In particular, after the application of the affine function $F$, an additional nonlinear function $G: \mathbb{R} ^ 2 \mapsto \mathbb{R} ^ 2$, can be applied to the coordinates. The latter is referred to as variation. The authors of \cite{draves1} provide 49 such variations, e.g:

\begin{figure*}[!t]
    \includegraphics[width=\linewidth]{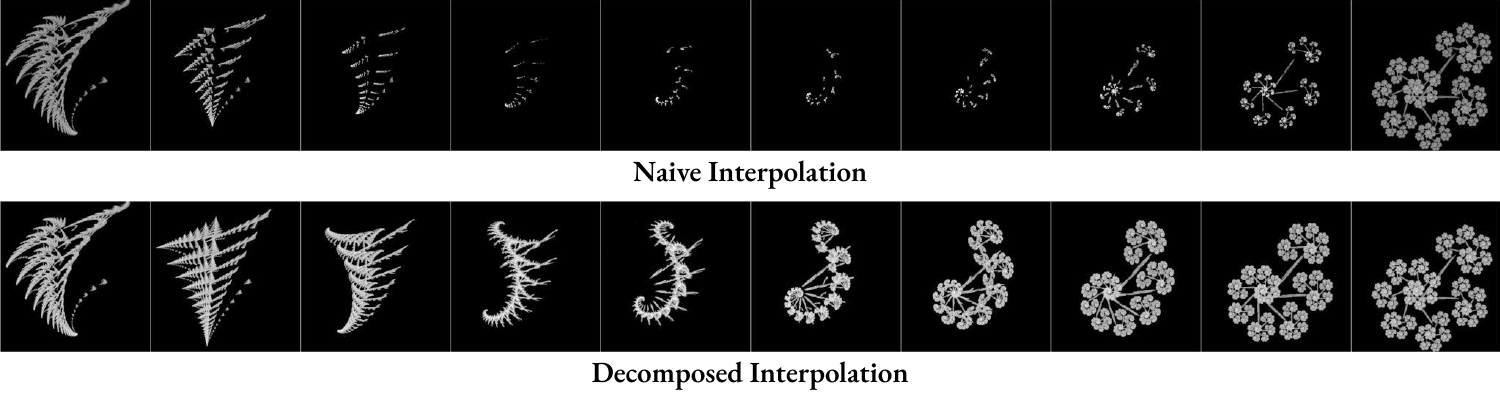}
    \caption{An example of the the proposed animation method compared to naive interpolation (Sec. \ref{sub:fractal_animation}). The latter often results in undesired sparseness in the intermediate frames. The former mitigates this issue. More samples are displayed in Fig. \ref{fig:appendix_fractal_video_linear} and \ref{fig:appendix_fractal_video_nonlinear} of Appendix \ref{appendix:sup_visual_material}.}
    \label{fig:fractal_video}
\end{figure*}

\begin{itemizetight}

\item $G_6(x, y) = r  (\sin(\theta + r)$
\item $G_{16}(x, y) = \frac{2}{r + 1} (y, x)$
\end{itemizetight}

Here $r$ and $\theta$ are polar coordinates. As seen in Fig. \ref{fig:attractor}, such images differ significantly from the ordinary IFS. Thus, the inclusion of nonlinear functions significantly boosts the overall diversity of generated samples. At the time of writing this document, no other work has explored fractal flames in the context of deep learning.

\subsection{Synthetic Videos based on Fractal Geometry}
\label{sub:fractal_animation}

Animation can be simply achieved by sampling parameters for two fractal images, the first and last frame. The only constraint is the number of functions $N$, which must be shared. To produce motion, parameters of the two images are linearly interpolated. As the order of functions in an IFS is arbitrary, we sort them by their probabilities $p_i$. The result is a smooth sequence of $T$ IFS images which can be rendered separately to produce a video. Nonetheless, although the beginning and end of the resultant clips are satisfactory, the intermediate attractors are not,  often exhibiting evident sparseness (Fig. \ref{fig:fractal_video}). We consider this behavior inappropriate for producing large-scale synthetic datasets. Therefore, an alternative solution is required.

Instead of directly interpolating IFS parameters, we propose to alleviate detrimental sparseness by employing matrix decomposition (Sec. \ref{subsub:classic_ifs}). Specifically, each parameter matrix $A_i$ can be decomposed into sub-matrices. The approach is to first interpolate each pair of sub-matrices separately and subsequently multiply them to obtain the final parameters of each frame. This method is adapted from \cite{anderson1} and \cite{burch1} and examples are displayed in Fig. \ref{fig:fractal_video}. The complete procedure is described in detail in Algorithm \ref{alg:decomposed_animation}. Regarding notation, $\texttt{zeros}(x)$ is a matrix of shape $x$ filled with zeros, $\texttt{diag}(x, y)$ is a diagonal $2D$ matrix with $x, y$ as elements, $\texttt{interp}(x, y, z)$ denotes linear interpolation between $x$ and $y$ of length $z$, $\texttt{rot\_matrix}(x)$ is a $2D$ rotation matrix parameterized by angle $x$ and the symbols $:$ and $\ldots$ have the same functionality as in \texttt{numpy}.

It is noteworthy that fractal geometry can generate videos  in an alternative manner: by constructing a point cloud with a 3D IFS. Specifically, by extracting  2D slices at different coordinates within this point cloud, a sequence of images can be produced. However, this approach is not pursued due to its higher computational demands and incompatibility with certain key modifications introduced in Sec. \ref{sub:domain_adaptation}.

\begin{algorithm}
\caption{\texttt{sample-video-decomposed}$()$: Sample IFS parameters for an animation by interpolating each sub-matrix separately.}
\label{alg:decomposed_animation}

\hspace*{\algorithmicindent} \textbf{Output}: Sequence of parameters: $W \in \mathbb{R} ^ {T \times N \times 6}$

\begin{algorithmic}[1]
\algrule

\State Sample $N  \thicksim U(\{3, \ldots, 8  \}) $ \Comment{\# functions}

\State Sample $T  \thicksim U(\{18, \ldots, 20  \}) $ \Comment{\# frames}

\algrule

\State $D \gets \texttt{zeros}(N, 2, 2) \in \mathbb{Z} ^ {N \times 2 \times 2}$ \Comment{Initialize matrices}

\State $\Sigma \gets \texttt{zeros}(T, N, 2, 2) \in \mathbb{R} ^ {T \times N \times 2 \times 2}$

\State $R_{\theta} \gets  \texttt{zeros}(T, N, 2, 2) \in \mathbb{R} ^ {T \times N \times 2 \times 2}$

\State $R_{\phi} \gets  \texttt{zeros}(T, N, 2, 2) \in \mathbb{R} ^ {T \times N \times 2 \times 2}$

\State $b \gets  \texttt{zeros}(T, N, 2) \in \mathbb{R} ^ {T \times N \times 2}$

\algrule

\For{$n = 1$ to $N$} 
    \State \texttt{Sample} $d_1, d_2 \thicksim U\left(\{-1, 1\}\right)$ 
    
    \State $D[n, \ldots] \gets \text{\texttt{diag}}(d_1, d_2) \in \mathbb{Z} ^ {2 \times 2}$

    \algrule

    \State \texttt{Sample} $\sigma_1^1, \sigma_2^1$ \Comment{see Appendix A of \cite{anderson1}}
    
    \State \texttt{Sample} $\sigma_1^T, \sigma_2^T$ 

    \State $\Sigma^1, \Sigma^T \gets \text{\texttt{diag}}(\sigma_1^1, \sigma_2^1), \text{\texttt{diag}}(\sigma_1^T, \sigma_2^T) \in \mathbb{R} ^ {2 \times 2}$

    \State $\Sigma[:, n, \ldots] \gets \text{\texttt{interp}} \left( \Sigma^1, \Sigma^T, T \right) \in \mathbb{R} ^ {T \times 2 \times 2}$
    
    \algrule

    \State \texttt{Sample} $\theta^1, \theta^T, \phi^1, \phi^T \thicksim U(0, 2\pi)$ 

    \State $\theta, \phi \gets \text{\texttt{interp}}(\theta^1, \theta^T, T), \text{\texttt{interp}}(\phi^1, \phi^T, T)$

    \For{$t = 1$ to $T$}
        \State $\R_\theta[t, n, \ldots] \gets \text{\texttt{rot\_matrix}} \left( \theta[t] \right) \in \mathbb{R} ^ {2 \times 2}$

        \State $\R_\phi[t, n, \ldots] \gets \text{\texttt{rot\_matrix}} \left( \phi[t] \right) \in \mathbb{R} ^ {2 \times 2}$

    \EndFor

    \algrule

    \State \texttt{Sample} $b^1, b^T \thicksim U\left(-1, 1\right) \in \mathbb{R} ^ {2}$ 

    \State $b[:, n, :] \gets \text{\texttt{interp}}(b^1, b^T, T) \in \mathbb{R} ^ {T \times 2}$

\EndFor

    \algrule

\State $A \gets R_{\theta} \Sigma R_{\phi} D \in \mathbb{R} ^ {T \times N \times 2 \times 2}$ 
\Comment{Compose $A$}

\State $W \gets \texttt{reshape}(\texttt{concat}(A, \texttt{expand}(b))) \in \mathbb{R} ^ {T \times N \times 6}$ 
\State \textbf{return} $W$    

\end{algorithmic}
\end{algorithm}

\subsection{Domain Adaptation}
\label{sub:domain_adaptation}

\begin{figure}
    \centering
    \includegraphics[width=\linewidth]{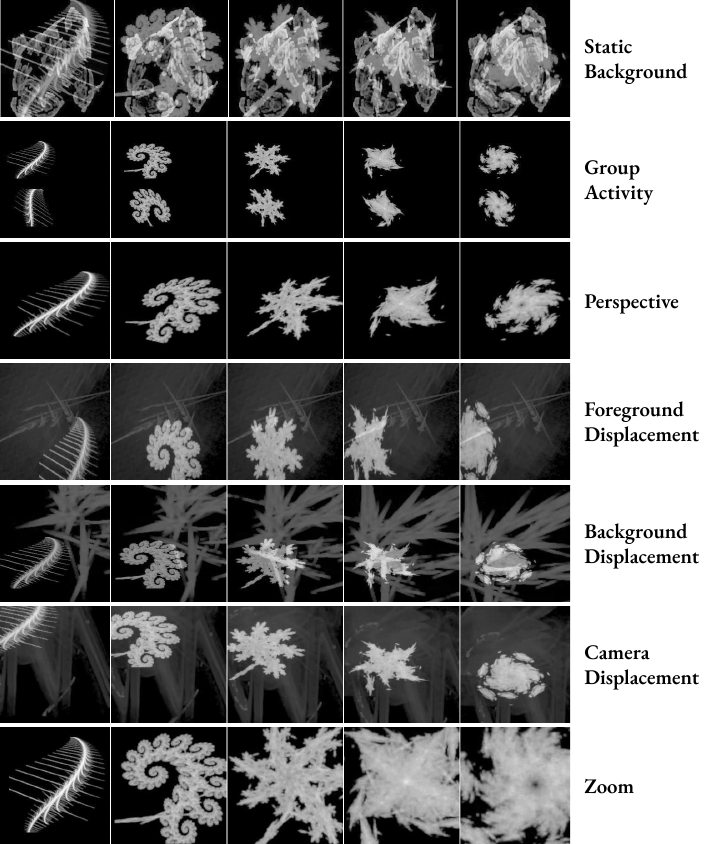}
    \caption{Examples of the proposed domain adaptation methods. The purpose of these augmentations is to narrow the domain gap between real and synthetic videos.}
    \label{fig:augmentations}
\end{figure}

Previous work on images \cite{baradad1} concludes that downstream performance is boosted if synthetic and real data share structural properties. Likewise, large-scale studies on pre-training \cite{cole1,kotar1,thoker1} deduce that its effectiveness significantly deteriorates if the source and target domains differ. As such, it can be assumed that obtaining satisfactory video models requires narrowing the domain gap between synthetic videos and samples from  action recognition benchmarks. To do so, this section lists manually observed characteristics of real action recognition data \cite{soomro1,kuehne1} as well as methods to emulate them within the synthetic pre-training framework (Fig. \ref{fig:augmentations}). Amongst these techniques, nonlinear motion and amplified diversity can only be applied offline, requiring the construction of a new video with modifications to Alg. \ref{alg:decomposed_animation}. On the contrary, the rest are implemented as online augmentations, further promoting the diversity of the generated videos on-the-fly.
It should be noted that in the context of our work, domain adaptation refers to the proposed set of heuristic augmentations which simulate mostly motion-related properties of real videos.

\begin{figure}
    \centering
    \includegraphics[width=\linewidth]{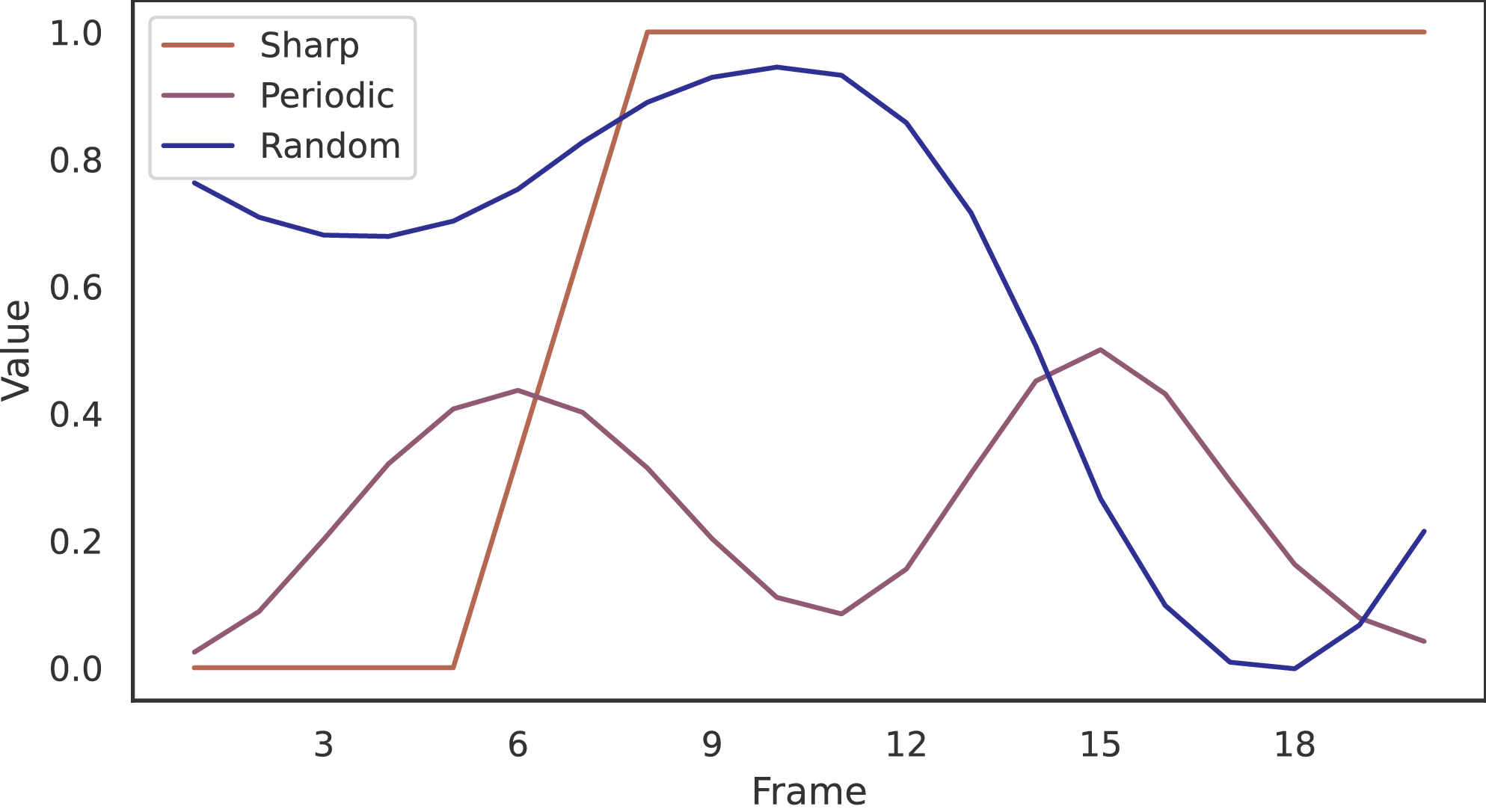}
    \caption{Examples of the proposed nonlinear interpolation curves. The objective is to approximate the complexity of real human motion.}
    \label{fig:nonlinear_interpolants}
\end{figure}

\textbf{Nonlinear Motion}. Our synthesis method produces simple forward motion. However, the real human counterpart is more complex. To this end, more intricate interpolation functions (Fig. \ref{fig:nonlinear_interpolants}) can be included in the rendering process (Alg. \ref{alg:decomposed_animation}):

\begin{itemizetight}
    \item \textbf{Sinusoidal Interpolation}. Periodic activity such as exercise can be simulated with a noisy sine  function. 
    
    \item \textbf{Sharp Interpolation}. Quick and sudden activity such as boxing can be approximated by a linear interpolant with a significantly sharper slope. The linear interpolant is placed in random timestep while the rest of the curve is padded. 
    
    \item \textbf{Random Interpolation}. Other activity without clear patterns can be simulated with a random waveform. To produce such a waveform, a sequence of real numbers is initially sampled from $U(0, 1)$ and then smoothened via 1D quadratic interpolation.
    
\end{itemizetight}

Moreover, human motion is inherently composite, i.e. intricate motions consist of multiple simple ones. For instance, running is comprised of a periodic movement of the legs as well as a different periodic movement of the arms. This can be approximated by assigning a different interpolant to each IFS function pair. As a result, the produced shape will execute multiple motions simultaneously. However, doing so unrestrictedly often results in oscillations that lack the structured flow found in real human motion. To this end, interpolation functions will be sampled under constraints. Initially, the set of chosen interpolants is initialized as the linear interpolant. Next, one or two different nonlinear interpolants are added to the set. Lastly, each of the $N$ IFS function pairs receives a randomly sampled interpolant from the set or a single interpolant is applied to all IFS functions. The $N$ resultant interpolants replace the \texttt{interp} operation in Alg. \ref{alg:decomposed_animation}.The outcome is a shape that moves in more fluid and constrained manner compared to naive sampling of interpolants.

\textbf{Diversity}. It has been demonstrated that diversity of synthetic images during pre-training leads to stronger visual  representations \cite{baradad1}. In the context of fractal animations, this property can be boosted by adding  nonlinearity to the image rendering algorithm (Section \ref{sub:fractal_flame}). In pursuit of reducing the domain gap, only variations $4, 14, 16, 17, 20, 27 \allowbreak \text{ and }  \allowbreak 29$ (Appendix of \cite{draves1}) are selectively employed. These were chosen due to  their  distinct and well-defined shape and contours, a characteristic present in real videos.

\textbf{Random Background}. Real videos consist of  two essential components: foreground (person performing an action) and background (environment surrounding the person). In its simplest form, the background is completely static. Although in synthetic videos this property is absent, it can be straightforwardly approximated following \cite{wang1,ding1}. For each video $x_i$ within a batch, a static frame is sampled from a different video $x_j$ and mixed with every frame of $x_i$ via weighted sum: 

\begin{align*}
    \Tilde{x_i} = (1 - a) x_i + a x_j[f]
\end{align*}

Here, $f \thicksim U(\{1, \ldots , \allowbreak T\}$ and $a \thicksim U(0.25, 0.55)$. In practise, to cover the entirety of the canvas,  $N_{back} = 2$ static frames are sampled and are next aggregated with the maximum  operation. Furthermore, a random rectangle is cropped from the resultant image and interpolated to input dimensions.

It has been assumed that the background remains motionless. However, real videos often include dynamic elements such as bystanders or water waves. This can be addressed with a modification to the previous approach. Specifically, the given video is mixed not with a single static frame but with a sequence of frames sampled from a different video. Compared to foreground, the magnitude of the background motion should be smaller. Thus, within this sequence, the frame index can either be incremented by one, remain the same, or decrease by one at each timestep. Additionally, the difference between the maximum and minimum frame indices is constrained. For each video in a batch, the type of background is determined by a Bernoulli trial with probability 0.8. Success results in static background, whereas failure in dynamic.

\textbf{Foreground Scaling and Placement}. In synthetic videos, fractal shapes cover the majority of the canvas and are usually positioned around its center. On the contrary, in real videos, position and size of the foreground significantly vary. To address this contradiction, synthetic videos are downsampled in the two spatial dimensions with scales $s_h, s_w \thicksim U(s_{min}, s_{max})$ and placed in a random position of an empty canvas. We set $s_{min} = 0.3$ and $s_{max} = 1.0$.

\textbf{Group Activity}. Videos that display groups performing similar activities (e.g. aerobics) can be approximated with a modification to the previous augmentation. Specifically, after the interpolation step, the synthetic video is copied resulting in  $N_{clone} = 2$ clones with each one receiving a different mild augmentation. Augmentations, which render the copies asynchronous, include random rotation, horizontal flipping and temporal offset. As previously, each copy is then placed in a random location of the canvas. We set $s_{min} = 0.2$ and $s_{max} = 0.7$ to reduce overlap between copies.

\textbf{Perspective}. Real cameras record objects  from any angle in 3D, whereas fractals are rendered in 2D. As a compromise,  minor angle variance can  be induced using a random perspective transformation\footnote{We employ the \texttt{RandomPerspective} transform from \texttt{torchvision}.}. This augmentation is  expected to amplify the model's spatial perception.

\textbf{Relative Displacement}. Aside from the previously mentioned motion, real videos contain additional relative motion between the foreground, the background and the camera:

\begin{itemizetight}
    \item \textbf{Foreground Displacement}. This occurs when camera is static and individuals perform actions while simultaneously walking or running. As such, in the captured video, the position of the foreground is shifted while the background remains unaffected.
    
    \item \textbf{Background Displacement}. This is observed when the human target is displaced and the camera follows it. Consequently, the position of the human subject  remains virtually stationary, while the background undergoes an equal displacement in the opposite direction. A notable example is the camera that follows athletes in a running track.
    
    \item \textbf{Camera Displacement}. In this case, the position of the human target remains unchanged but the focus of the camera is being shifted. In the resultant video, both foreground and background are relatively displaced in the opposite direction of the camera's movement.
\end{itemizetight}

Invariance to such movements can be boosted with simple transformations. For background displacement, a static background frame is initially enlarged and then a sequence of crops with dimensions of the original video is created. The sequence is then mixed with the foreground. As the centers of the crops are consecutive points on a 2D line, the result is displacement towards a fixed direction. For camera displacement, the process is similar, with the exception that crops are taken after mixing the foreground with an unaltered background. Lastly, for foreground displacement, the foreground is initially reduced in size and then each frame is placed at a different position inside the background.

\textbf{Camera Zoom}. For implementation, the video is initially interpolated to larger spatial dimensions. This is followed by central cropping which is applied with different scale for each frame. Increasing and decreasing the scale results in zooming out and zooming in respectively. As a final step, each cropped frame is interpolated back to the original spatial dimensions of the video. This implementation differs from the previously proposed camera displacement. The latter alters the position the displayed shapes between consecutive frames. The former alters their size.

\begin{figure}
    \centering
    \includegraphics[width=\linewidth]{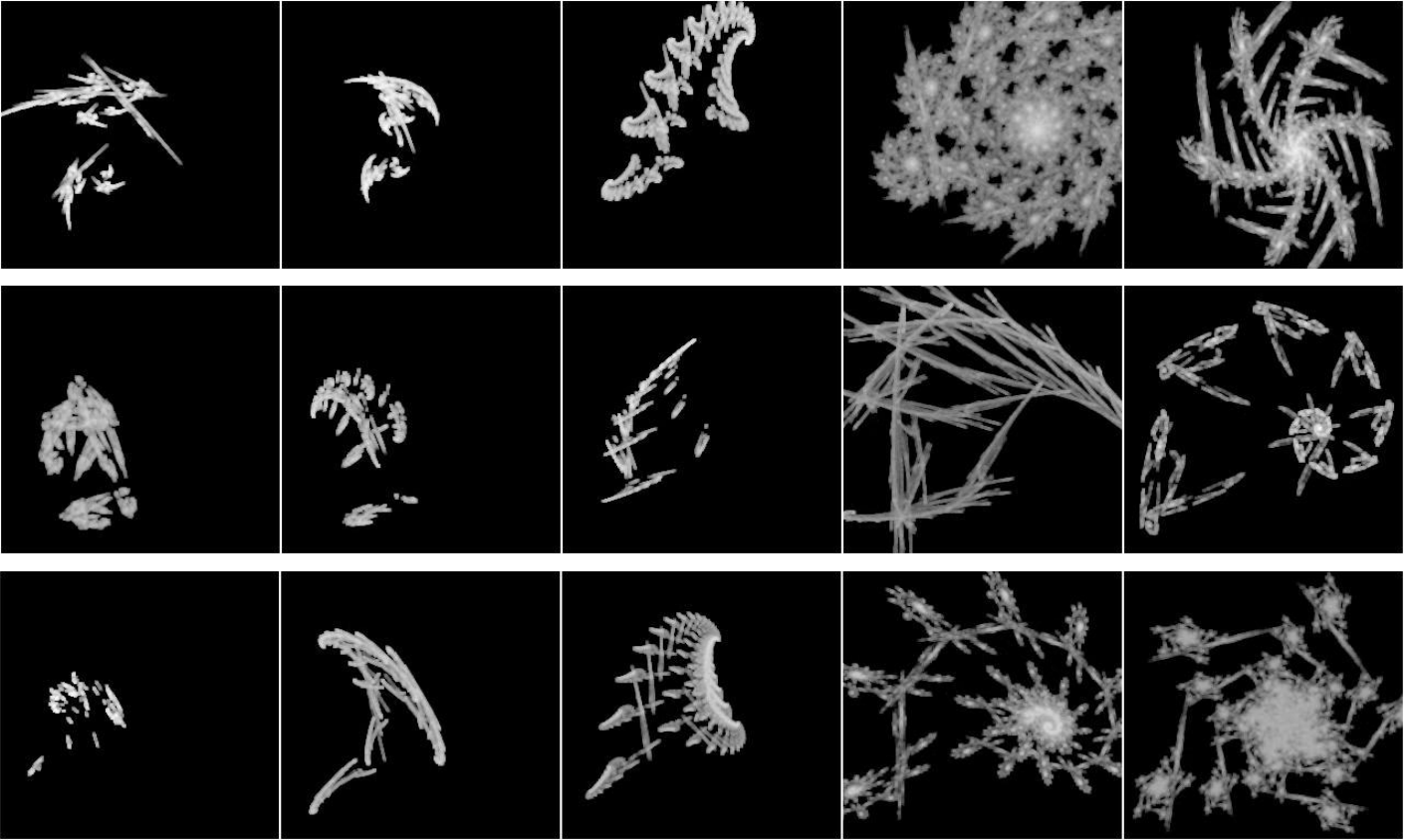}
    \caption{Examples of the proposed mutation scheme. Each row displays a separate video. All videos belong to the same category, but exhibit differences due to randomly sampled noise, which is injected into the parameters.}
    \label{fig:mutations}

\end{figure}

\textbf{Camera Shake}. Videos captured by hand-held devices often contain shaking. This phenomenon can be approximately synthesized following \cite{qu1}. Specifically, the displacement in each dimension is modelled as: 

\begin{align*}
    d_t = \sum_{i=1}^{n} \frac{1}{i} \sin (2 \pi f_i t + \phi) + \eta_t, t = 1, 2, \ldots , T
\end{align*}

Here, $n$ is the number of components, $T$ is the duration of the video in frames, $f_i$ is the frequency of each component, $\phi$ denotes the phase and $\eta$ is the noise component. These parameters are sampled as follows: $n \thicksim U(\{2, \ldots 5 \}), f_i \thicksim U(0.1, 1.2), \phi_i \thicksim U(0, 2 \pi), \eta_t \thicksim U(-0.3, 0.3)$. To apply the shaking effect, two displacement sequences are sampled. Afterwards, the video is enlarged and each frame is cropped. The position of the crop is  determined by the two displacement values. This operation can be considered as an extension of the camera displacement augmentation, as it replaces the simple forward camera motion with a complex oscillation.

\subsubsection{Automatic Construction of Categories}
\label{sub:automatic_category}

Algorithm \ref{alg:decomposed_animation} describes a method to sample parameters for a fractal video. 
Repeatedly executing this algorithm results in a dataset where each video is unique and no information exists about correlation of different samples.  Therefore, in the context of machine learning, such a dataset is only suitable for unsupervised learning. 

For a supervised classification objective, synthetic videos must be divided into categories (Fig. \ref{fig:mutations}). These can be automatically constructed by adapting the approach of \cite{kataoka1}. Specifically, given a predefined number of categories $C$, we initially sample parameters for $C$ distinct fractal videos. This is achieved by executing Alg. \ref{alg:decomposed_animation} modified with nonlinear motion and diversity domain adaptations (Sec. \ref{sub:domain_adaptation}). As such, each category $c$ is represented by a parameter matrix $W_c \in \mathbb{R} ^ {T_c \times N_c \times 6}$ as well as a variation index  $var_c$. Here, $T_c$ is the number of frames in the video, $N_c$ the number of IFS functions and $var_c$ determines the type of nonlinear function utilized in rendering the video (Sec. \ref{sub:fractal_flame}).

To produce a new video belonging to class $c$, we mutate the respective parameters $w_c$ with the auxiliary ``noise" matrices $m_{a}$ and $m_{b}$. The parameter matrix of the final video is calculated as follows: 

\begin{align*}
    \Tilde{W_c} = m_{a} \odot W_c + m_{b}
\end{align*}

Here, $m_{a} \in \mathbb{R} ^ {T_c \times 1 \times 6}$ is the scaling component of the noise, which consists of $6$ random curves that are sampled in the same manner as the random nonlinear interpolant (Sec. \ref{sub:domain_adaptation}). 
The curves are bound between $[-0.35, 0.35]$.
On the other hand, $m_{b} \in \mathbb{R} ^ {1 \times N_c \times 6}$ is the bias component, which is sampled from $U(-0.2, 0.2)$. 
Lastly, $\odot$ denotes elementwise multiplication. With this mutation scheme, we achieve variance within the same class and increased difficulty during supervised pre-training. However, it is noteworthy that the produced categories are randomly sampled and therefore, unlike real datasets such as Kinetics \cite{carreira2,kay1}, possess no interpretable information (i.e., each class can not represent an actual human action).

\begin{figure}
    \centering
    \includegraphics[width=\linewidth]{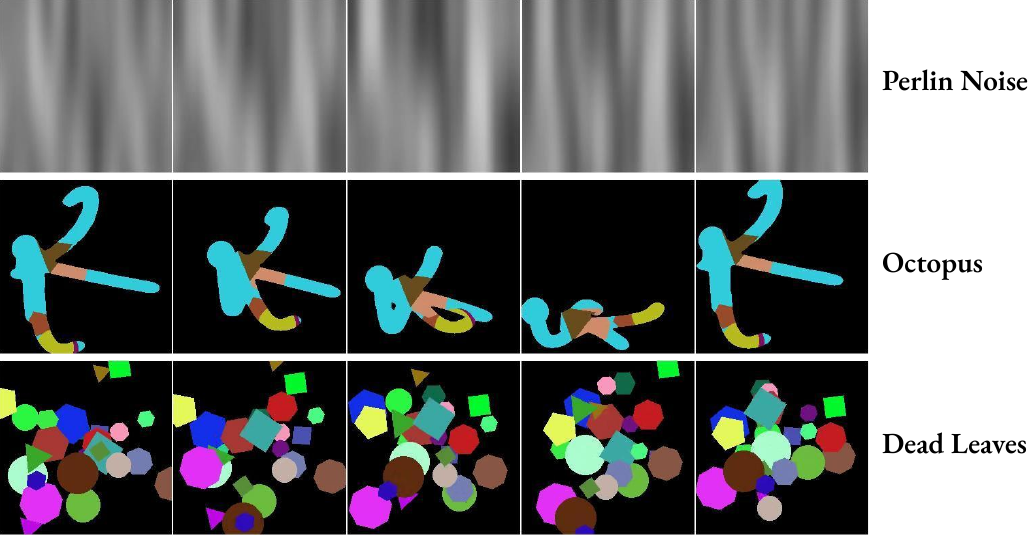}
    \caption{Examples of alternative synthetic videos. Compared to fractals, these videos are less diverse, whereas perlin noise and dead leaves additionally lack distinct contours.}
    \label{fig:alternative_videos}
\end{figure}

\subsection{Alternative Synthetic Videos}
\label{sec:alternative_video}

Despite their appealing properties (Sec. \ref{sub:fractal_intro}), it is not evident that fractal animations are appropriate for training strong visual representations for the task of action recognition. Hence, this section presents alternative generative processes of video (Fig. \ref{fig:alternative_videos}), which will be compared against fractals during  experiments. Each generative process produces videos with different characteristics and the objective of the upcoming experiments is to determine which attributes are favorable for downstream results.

\textbf{Perlin Noise}. Adapting the approach of \cite{kataoka3}, we can generate videos of perlin noise \cite{perlin1,perlin2}, a variant of random texture. The label of each video is defined by three frequencies: two spatial and one temporal. These determine the rate of change in their respective dimensions. Unlike fractals, these videos are nebulous, lacking distinct shape and contours. With regard to the proposed domain adaptation techniques, perlin noise is not compatible with diversity and nonlinear motion.

\textbf{Octopus}. By sampling two 1D waveforms, one can construct a random 2D curve. To create an animation, two such curves can be sampled and their coordinates interpolated as has been shown for IFS. This process is repeated $N$ times and the resulting curves are conjoined at a fixed point. As the outcome consists of thin curves, thickness is increased by applying Gaussian blur  followed by the operation of morphological closing \cite{vincent1}. These videos will be referred to as \say{octopus} due to similarity to the mollusk. We additionally apply colorization, interior removal via the morphological gradient \cite{vincent1} and addition of geometrical shapes. Octopus videos resemble fractals due to their distinct contours, but are less diverse. These videos are not suitable for the proposed diversity amplification technique.

\textbf{Dead Leaves}. Dead leaves \cite{ruderman1,lee1} is a simple image model that emulates statistics of natural images, such as the $1/|f| ^ a$  power spectrum. Such images can be constructed by randomly filling a canvas with geometric shapes, such as circles and regular polygons. More recently, dead Leaves have been employed in deep learning as synthetic images for pre-training \cite{baradad1} and it was deduced that  stronger representations are obtained when the said shapes vary in terms of size, color and number edges. 
To extend this generative process to the video domain, a 2D curve is sampled for each shape. Throughout the video, the shape traverses this curve. Lastly, dead leaves lack distinct contours, while being incompatible with  diversity enhancement.

\subsection{Training Objective}\label{sub:training_objective}

So far, we have described the proposed method and the respective training procedure. 
However, the proposed training relies on supervised learning over a pre-defined number of categories, arbitrary generated by our sampling procedure.
Despite the non-intuitive formalization of classes (non-intuitive in the sense that the defined distinct categories are arbitrarily selected and do not correspond to a human-related action), such a supervised learning approach is expected to provide good visual embeddings. 
On the other hand, one may wonder what happens if we do not define such distinct classes and treat our problem as a self-supervised paradigm. 
To this end, for pre-training, aside from the supervised objective from Sec. \ref{sub:automatic_category}, we additionally explore algorithms from the self-supervised learning  (SSL) literature, which do not require annotation.  During SSL training, a pretext task is designed for a deep learning algorithm to solve and pseudolabels for the pretext task are automatically constructed based on  attributes of the input. 
Specifically we employ the SSL frameworks SimCLR \cite{chen3}, MoCoV2 \cite{chen1} and BYOL \cite{grill1}.

\textbf{SimCLR}. Given a batch of size $B$,
SimCLR applies heavy augmentation twice resulting in  $2B$ input samples. For each positive pair $(i, j)$ originating from the same sample,  the other  $2(B - 1)$ instances are treated as negatives. The contrastive prediction loss  is calculated as:

\begin{align*}
    L_{i,j} = - \log \frac {\exp(\mathrm{sim}(z_i, z_j )/T ) } {\sum_{\substack{m=1 \\ m\neq i}}^{2B}\exp(\mathrm{sim}(z_i, z_m)/T )}
\end{align*}

Here, $z_i$ is the model output for the instances $i$, $\mathrm{sim}(\makebox[1ex]{\textbf{$\cdot$}} , \makebox[1ex]{\textbf{$\cdot$}})$ is cosine similarity and  $T$ is the temperature hyperparameter. SimCLR requires two forward and two backward passes per training step. Furthermore, large batch size is preferred for training as it leads to increased difficulty and improved downstream results. As such, SimCLR is computationally heavier than supervised classification.

\textbf{MoCoV2}. MoCoV2 similarly employs a contrastive approach resulting in two views per input: $q$ and $k_+$. However, unlike before, $k_+$ is produced by a non-differentiable model which is updated as an exponential moving average of the original. Moreover, $k_+$ is added to circular queue of size $K$. The loss function maximizes the similarity between $q$ and $k_+$ while minimizing it between $q$ and all other instances in the queue:

\begin{align*}
    L = - \log  \frac {\exp(q \cdot k_+/T )} {\sum_{i=1}^{K}{\exp(q \cdot k_i/T)}}
\end{align*}

In terms of resources, MoCoV2 requires two forward and one backward pass, while the dependency on large batch size is alleviated due to the queue. Hence MoCoV2 is more lightweight than SimCLR but still more demanding than supervised classification.

\textbf{BYOL}. Lastly, BYOL does not utilize  negative pairs. In a manner akin to MoCoV2, BYOL employs two neural networks named online and target, with only the former one being differentiable. As before, two views of the input are constructed: $q$ and $k$. Both views are fed through both encoders resulting in four output representations in total. The  training objective is to simply maximize the cosine similarity between all matching representations:

\begin{align*}
    & q_{o}, k_{o} = \mathrm{encoder\_online}(q), \mathrm{encoder\_online}(k) \\
    & q_{t}, k_{t} = \mathrm{encoder\_target}(k),  \mathrm{encoder\_target}(q) \\
    & L = 2 - 2 * \left(\mathrm{sim}(q_{o}, k_{t}) + \mathrm{sim}(k_{o}, q_{t}) \right)
\end{align*}

Computationally, each training step requires four forward and two backward passes, rendering it the most demanding framework in the present work. Nonetheless, unlike SimCLR, large batch sizes are not required.

%% file: 3-experiments.tex
\begin{figure*}[t]
    \includegraphics[width=\linewidth]{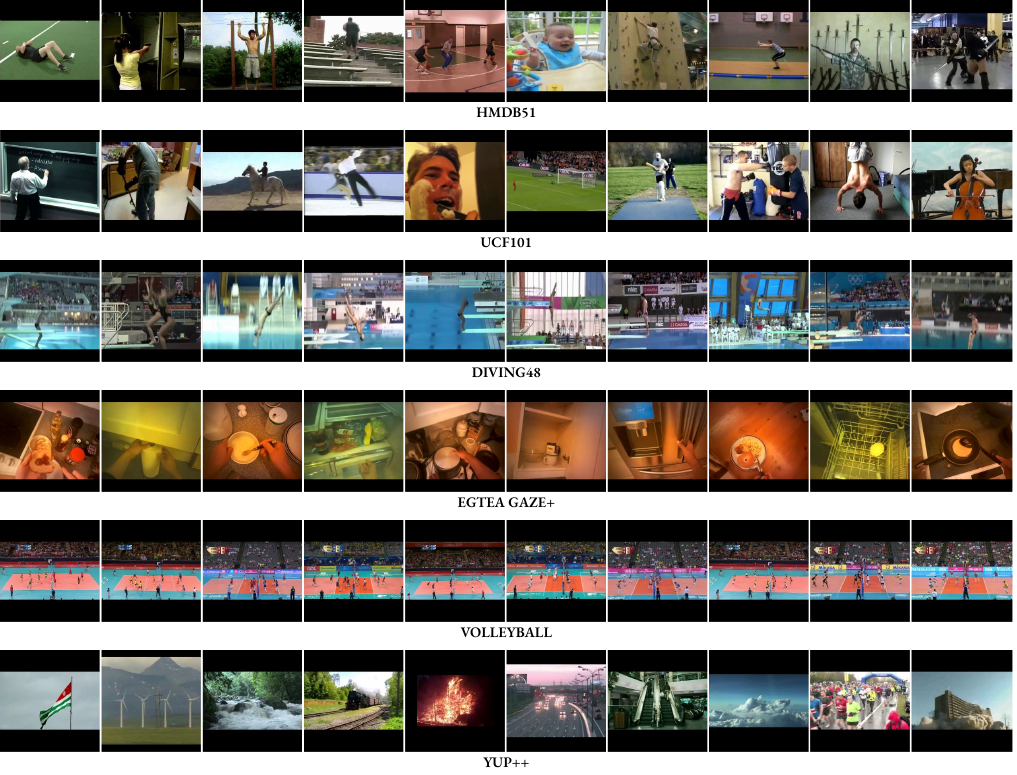}
    \caption{Frames taken from the evaluated downstream datasets. Note the variance of the displayed background in all datasets aside from VOLLEYBALL and DIVING48.} \label{fig:downstream}
\end{figure*}

\section{Experiments}
\label{sec:experiments}

\subsection{Implementation Details}

\textbf{Datasets}. The proposed pre-training framework is evaluated by fine-tuning the model on 6 small-scale video classification datasets:

\begin{itemizetight}
    \item HMDB51 \cite{kuehne1} and UCF101 \cite{soomro1}, established action recognition benchmarks.

    \item DIVING48 \cite{li1}, a collection of videos from diving competitions.

    \item EGTEA GAZE+ \cite{li2}, which consists of first person cooking videos.

    \item VOLLEYBALL \cite{ibrahim1}, a group action recognition dataset.

    \item YUP++ \cite{feichtenhofer1}, which consists of videos depicting dynamic scenes that are not relevant to action recognition.
\end{itemizetight}

Indicative examples from these datasets are displayed in Fig. \ref{fig:downstream}. The datasets exhibit significant differences and were chosen to maximize overall diversity. An official train-validation split is provided for each dataset. Thus, after pre-training, we fine-tune models on the training set and report the top-1 accuracy on the validation set of the downstream datasets.
Synthetic videos are rendered with spatial resolution of $256 \times 256$ and temporal length which is sampled from U(\{18, \ldots,  20\}).
For fractals, 50\% of videos are rendered with nonlinearities (Sec. \ref{sub:fractal_flame}). 

\textbf{Model Architecture}. Temporal Shift Module (TSM) \cite{lin1} is employed for all experiments with ResNet-50 \cite{he1} utilized as backbone. While preserving the efficiency of 2D CNNs, TSM can achieve results equivalent to 3D CNNs in action recognition tasks.  Specifically, TSM efficiently exchanges information between neighboring frames by moving the feature map along the temporal dimension. Despite being computationally cheap, this operation possesses a strong spatio-temporal modeling ability. Solid accuracy can be achieved with as few as $8$ input frames. Such short length significantly alleviates both the CPU and GPU bottlenecks. The former is evidenced by reduced dataloading whereas the latter by a reduced amount of computation within the neural network itself.  Unless specified otherwise, pre-training is done from scratch and does not utilize any off-the-shelf checkpoints.

\begin{table*}
\begin{center}
\begin{tabular}{l|c|cccccc} \hline
    Method & Kept & HMDB51 & UCF101 & DIVING48 & EGTEA & VOLLEY & YUP \\ \hline\hline

    Scratch & - & 31.5  & 70.3  & 4.7 & 50.5 & 53.8 & 43.7 \\ \hline
 
    Initial & - & 41.4 & 72.7 & 24.3 & 49.8 & 80.6 & 52.3 \\ \hline
    
    Background & Yes & \cellcolor{green!50}47.4 & \cellcolor{green!50}74.2 & \cellcolor{red!50}23.0 & \cellcolor{red!50}47.1 & \cellcolor{red!50}77.6 & \cellcolor{green!50}56.4\\ \hline
    
    Motion & Yes & \cellcolor{orange!50}47.3 & \cellcolor{green!50}75.7 & \cellcolor{red!50}21.7 & \cellcolor{green!50}48.6 & \cellcolor{green!50}79.2 & \cellcolor{green!50}59.5 \\ \hline
    
    Diversity & Yes & \cellcolor{blue!50}50.4 & \cellcolor{blue!50}77.8 & \cellcolor{blue!50}24.9 & \cellcolor{blue!50}49.8 & \cellcolor{blue!50}80.8 & \cellcolor{blue!50}63.1\\ \hline
    
    Scale + Shake & Yes & \cellcolor{green!50}52.8 & \cellcolor{green!50}78.0 & \cellcolor{green!50}26.0 & \cellcolor{green!50}50.7 & \cellcolor{red!50}80.8 & \cellcolor{orange!50}61.9 \\ \hline
    
    Shift & Yes & \cellcolor{green!50}\textbf{54.5} & \cellcolor{green!50}79.6 & \cellcolor{green!50}29.5 & \cellcolor{green!50}50.8 & \cellcolor{green!50}80.9 & \cellcolor{green!50}65.1\\ \hline
    
    Zoom & Yes & \cellcolor{orange!50}54.3 & \cellcolor{green!50}\textbf{80.2} & \cellcolor{green!50}\textbf{30.8} & \cellcolor{green!50}\textbf{52.0} & \cellcolor{green!50}81.4 & \cellcolor{green!50}65.2 \\ \hline
    
    Perspective & No & \cellcolor{yellow!50}53.3 & \cellcolor{yellow!50}78.7 & \cellcolor{yellow!50}25.0 & \cellcolor{yellow!50}50.0 & \cellcolor{yellow!50}80.7 & \cellcolor{yellow!50}62.6\\ \hline
    
    Group & No & \cellcolor{orange!50}53.3  & \cellcolor{orange!50}79.7 & \cellcolor{orange!50}30.3 & \cellcolor{orange!50}51.9 & \cellcolor{green!50}\textbf{81.9} & \cellcolor{green!50}\textbf{66.0} \\ \hline

\end{tabular}
\end{center}
\caption{Experiment 1 - Effectiveness of domain adaptation. In the majority of cases, emulation of a property boosts accuracy if the downstream dataset includes it (\textcolor{green}{green color}) and deteriorates accuracy if it does not (\textcolor{red}{red color}).}
\label{tab:exp1-domain}
\end{table*}

\textbf{Training}. Only RGB frames are employed as model input in this work. Unless specified otherwise, the spatial resolution of the model input is $112 \times 112$. 
The clip consists of 8 strided frames. 
This allows us to decode only a fragment of the video file reducing dataloading latency. 
The stride of the input is 2 frames for pre-training, 4 for fine-tuning VOLLEYBALL and 6 for fine-tuning all other downstream datasets. 
During  each training epoch, one such clip is produced from every video by sampling the index of the first frame from the uniform distribution.
During validation, 10 such clips are uniformly selected from a video and separately fed into the model. The final output is the average of softmax scores.

The number of training epochs is 25 and 100 for pre-training and fine-tuning respectively. 
The number of warmup epochs for the learning rate scheduler is 3 and 10 respectively.  The networks are optimized using AdamW \cite{loshchilov1} with $\beta_1 = 0.9$ and $\beta_2 = 0.999$. Cosine scheduling \cite{loshchilov2} is employed.  The true learning rate is scaled according to the batch size: $lr_{true} = \frac{bs}{bs_{base}} lr$, where the batch size and base batch size are set to 32 and 32 respectively. The default learning rate is initialized at $1 \cdot 10 ^ {-6}$, increases to $8 \cdot 10 ^ {-4}$ during warmup and eventually falls to $1 \cdot 10 ^ {-5}$ at the end of training.  
The default weight decay is set to $1 \cdot 10 ^ {-2}$. Exceptions are the VOLLEYBALL dataset where the learning rates are $2.5 \cdot 10 ^ {-6}$, $2.5 \cdot 10 ^ {-3}$ and $2.5 \cdot 10 ^ {-5}$ and weight decay is set to $1 \cdot 10^{-1}$ as well as DIVING48 where  the learning rates are $1.5 \cdot 10 ^ {-6}$, $1.5 \cdot 10 ^ {-3}$ and $1.5 \cdot 10 ^ {-5}$.

\textbf{Augmentation}. For fine-tuning, augmentation consists of random cropping with bicubic interpolation, horizontal flip, Randaugment \cite{cubuk1} and Gaussian blur. For cropping, scale is sampled from $U(0.2, 1.0)$ and ratio from $U(0.75, 1.33)$. Augmentations are applied in the same order as they are listed. As an exception, for VOLLEYBALL we use area interpolation and omit horizontal flip and Gaussian blur. 

For pre-training, we retain the same augmentations with the addition of the proposed domain adaptation techniques. Within the aforementioned order, these are applied  between horizontal flip and Randaugment. We employ a curriculum and linearly increase the intensity of domain augmentations for the first 5 epochs.  
All domain augmentations are applied with a probability of 0.3 with the exception of Background, Scale, Perspective and Group whose probabilities are 1.0, 1.0, 0.8 and 0.15 respectively. To accelerate computation, each domain adaptation method is applied in parallel and identically to all of its selected samples inside a batch. Background Randomization, Scale and Group are exceptions that are applied independently for each sample.

\subsection{Experimental Results}

After each experiment, we retain the configuration with highest accuracy on HMDB51 and UCF101, the focus of our work. Highest accuracy is indicated with \textbf{bold} font in the provided tables.

\textbf{Experiment 1 - Domain Adaptation}. Here we evaluate the proposed domain adaptation techniques, which are inserted sequentially into the pre-training framework. A technique will be discarded if improvement is not achieved for HMDB51 and UCF101. All pre-training is done with the MoCoV2 \cite{chen1} self-supervised framework, due to its computational efficiency. The pre-training dataset consists of 100K unlabeled fractal videos.

From Table \ref{tab:exp1-domain} it can be observed that, aside from a few exceptions (\textcolor{orange}{orange color}), emulation of a property is beneficial for datasets that include it (\textcolor{green}{green color}) and detrimental for datasets that do not (\textcolor{red}{red color}). 
Hence, it can be inferred that self-supervised pre-training is more effective when the source and target domains are similar. As such, synthetic datasets should be maximally customized to mirror the properties of the target downstream dataset. This observation is in complete agreement with previous work \cite{cole1,kotar1,thoker1}. A notable example is background randomization which increases accuracy for HMDB51 and UCF101, but decreases it for DIVING48 and VOLLEYBALL. For the former, background is not relevant for the category of the video and varies in each sample. However, for the latter,  the background in all samples is similar and it can be assumed that neural models take it into account during classification.

Additionally, the only modification that results in non-trivial improvement across all benchmarks is amplified diversity of synthetic videos (\textcolor{blue}{blue color}). This is again in line with previous work on the image modality \cite{baradad1}. One the contrary, the only modification that results in overall deterioration is random perspective (\textcolor{yellow}{yellow color}). It is noteworthy that compared to training from scratch, considerable improvement in results is attained across all datasets except EGTEA. 
This shortcoming is further investigated in Sec. \ref{sec:manual_error_analysis}.

\textbf{Experiment 2 - Alternative Synthetic Data}. We investigate pre-training with alternative synthetic datasets from Section \ref{sub:automatic_category}. Each dataset exhibits different characteristics and the aim is to determine the ones that are favorable for downstream results.

\begin{table*}
\begin{center}
\begin{tabular}{l|cccccc} \hline
    Dataset & HMDB51 & UCF101 & DIVING48 & EGTEA & VOLLEY & YUP \\ \hline\hline

    Octopus & 50.9 & 76.5 & 25.5 & 49.1 & 80.7 & 58.3 \\ \hline

    Dead Leaves & 39.9 & 70.2 & 15.8 & 47.3 & 75.7 & 50.2 \\ \hline

    Perlin Noise & 42.4 & 72.5 & 21.4 & 50.1 & 76.4 & 58.2 \\ \hline

    Fractal & \textbf{54.3} & \textbf{80.2} & 3\textbf{0.8} & \textbf{52.0} & \textbf{81.4} & \textbf{65.2} \\ \hline
    
\end{tabular}
\end{center}
\caption{Experiment 2 - Evaluation of different synthetic datasets. Fractal videos outperform all alternatives. The characteristics that render them superior are diversity and distinct contours.} 
\label{tab:exp2-data}
\end{table*}

\begin{table*}
\begin{center}
\begin{tabular}{l|cccccc} \hline
    Objective & HMDB51 & UCF101 & DIVING48 & EGTEA & VOLLEY & YUP \\ \hline\hline

    MoCoV2 & 54.3 & 80.2 & 30.8 & 52.0 & 81.4 & 65.2 \\ \hline

    SimCLR & 57.5 & 81.3 & 34.3 & 54.7 & \textbf{83.2} & 70.2 \\ \hline

    BYOL & 52.0 & 78.1 & 26.1 & 52.0 & 80.4 & 66.0 \\ \hline

    Supervised & \textbf{61.5} & \textbf{84.9} & \textbf{38.5} & \textbf{54.8} & 82.7 & \textbf{73.6} \\ \hline

\end{tabular}
\end{center}
\caption{Experiment 3 - Exploration of different training objectives. Pre-training through  supervised classification yields superior results compared to SSL. A plausible  rationale is that supervision is more resistant to the domain gap between real and synthetic videos.}
\label{tab:exp3-objective}
\end{table*}

Based on Table \ref{tab:exp2-data}, fractal videos lead to superior results across all benchmarks compared to alternatives. Thus it can be deduced that fractals possess more appropriate properties for downstream tasks. Specifically, compared to the octopus dataset, fractals exhibit more diversity. On the other hand, what differentiates fractals from dead leaves and perlin noise is distinct contours. As distinct contours are an attribute of real videos, it is safe to assume that including them in a synthetic dataset bridges the domain gap. The importance of this trait is further highlighted by the fact that octopus videos outperform both perlin noise and dead leaves. As such, the verdict of this experiment is identical to the previous one: to obtain stronger visual representations, synthetic video datasets have to exhibit diversity and mimic characteristics present in target data.

\textbf{Experiment 3 - Training Objective}. So far, MoCoV2 \cite{chen1} has been exclusively employed for pre-training. This decision was made due to its computational efficiency. Thus, with the intention of maximizing downstream performance, this section explores alternative training objectives: self-supervised frameworks SimCLR \cite{chen3} and BYOL \cite{grill1} (Sec. \ref{sub:training_objective}) as well as a supervised classification objective (Sec. \ref{sub:automatic_category}). For the former, the unlabeled fractal dataset from previous experiments is reused. For the latter, a new dataset is constructed with 500 classes and 200 samples per class, resulting in 100K training videos in total.

As seen in Table \ref{tab:exp3-objective}, the supervised objective confidently outperforms alternatives. At first glance this is perhaps surprising as the categories are sampled randomly and therefore lack meaningful information present in real classification datasets. A possible explanation relies on the fact that self-supervised frameworks have been shown to be especially vulnerable to the domain gap  between source and target datasets \cite{cole1,kotar1,thoker1}. Therefore, it can be assumed that the supervised pre-training objective is more resilient to the difference in domains and therefore results in representations that are transferred more robustly to a wide range of downstream tasks.

On a different note, compared to supervised training, self-supervised counterparts require multiple forward passes per step and benefit from larger batch size and increased number of epochs. Hence, it is likely that our SSL models are undertrained and allocating more resources would considerably improve downstream results. Regardless, supervised training performs well under resource constraints and therefore is a more cost-effective solution. A last observation involving the examined SSL frameworks is that SimCLR outperforms MoCoV2, which in turn outperforms BYOL. This is in complete contrast with ImageNet pre-training, where the order is reversed \cite{grill1}.

\textbf{Experiment 4 - Dataset Size}. This segment investigates how transferability to downstream tasks is impacted by two statistical characteristics of the synthetic classification dataset: the number of classes and the number of instances per class. The experiment is divided into two stages. In the first stage, the number of classes is fixed to 500 as in the previous experiment, while the instances are varied at 100, 200, and 400 per class. In the second stage, the number of instances is fixed to the optimal value determined in the first stage, while  classes are varied at 250, 500 and 1000. To ensure fairness, each dataset is a superset for all smaller ones. For the largest of the assessed datasets, we additionally evaluate the perspective transform (Sec. \ref{sub:domain_adaptation}), which previously led to deteriorated results in Experiment 1.

\begin{table*}
\begin{center}
\begin{tabular}{l|cccccc} \hline
    \#Instance/\#Total & HMDB51 & UCF101 & DIVING48 & EGTEA & VOLLEY & YUP \\ \hline\hline

    100/50K & 56.5 & 81.8 & 35.8 & 52.6 & 82.6 & 69.3 \\ \hline
    
    200/100K & 61.5 & 84.9 & 38.5 & \textbf{54.8} & 82.7 & 73.6 \\ \hline

    400/200K & \textbf{61.5} & \textbf{86.0} & \textbf{40.8} & 53.8 & \textbf{83.0} & \textbf{75.1} \\ \hline

\end{tabular}
\end{center}
\caption{Experiment 4A - Impact of the number of instances per category on downstream accuracy.}
\label{tab:exp4-1-scale}
\end{table*}

\begin{table*}
\begin{center}
\begin{tabular}{l|cccccc} \hline
    \#Classes/\#Total & HMDB51 & UCF101 & DIVING48 & EGTEA & VOLLEY & YUP \\ \hline\hline

    250/100K & 60.3 & 85.3 & 38.1 & 53.5 & 84.1 & 75.6 \\ \hline

    500/200K & 61.5 & 86.0 & 40.8 & 53.8 & 83.0 & 75.1 \\ \hline

    1000/400K & 62.4 & 87.3 & \textbf{41.2} & \textbf{56.1} & \textbf{84.0} & \textbf{76.0} \\ \hline

    1000/400K + Perspective & \textbf{65.4} & \textbf{87.6} & 40.3 & 56.0 & 83.1 & 72.7 \\ \hline

\end{tabular}
\end{center}
\caption{Experiment 4B - Impact of the number of categories on downstream accuracy.}
\label{tab:exp4-2-scale}
\end{table*}

Observing Table \ref{tab:exp4-1-scale}, it can be deduced that downstream performance is almost a monotonically increasing function of the number of instances per class. Judging by Table \ref{tab:exp4-2-scale}, the same conclusion can be reached for the number of classes. This behavior is not surprising as increasing the number of training videos exposes the model to more spatio-temporal patterns and renders the learnt representations more transferable to new tasks.

Regarding the perspective augmentation, in Table \ref{tab:exp4-2-scale} accuracy is boosted for for action recognition datasets HMDB51 and UCF101. This is in contrast with Experiment 1 where performance drops. The difference between these experiments is the training objective: the former employs a supervised objective while the latter utilizes the self-supervised framework MoCoV2 \cite{chen1}. As such, it is possible that the exact impact of the proposed domain adaptation techniques is dependant on the training objective.

\textbf{Experiment 5 - Higher Resolution}. Given that all previous experiments have been conducted with a low spatial resolution of $112$, we now increase it to $224$. Additionally, we compare the results of fractal pre-training to the Kinetics counterpart \cite{carreira2,kay1}, which is the established  approach for pre-training in action recognition tasks. To this end, we employ an off-the-shelf checkpoint from \cite{lin1} that has also undergone training with a resolution of $224$.
Fine-tuning after Kinetics  utilizes different hyperparameters which are documented in Appendix \ref{appendix:kinetics_hyps}. It is noteworthy that Kinetics and fractal  pre-training are not equivalent. The former utilizes a smaller dataset of $250K$ training samples compared to the latter which uses $400K$. However, Kinetics training lasts for $100$ epochs in \cite{lin1}, whereas our experiments with fractals require only $25$.

\begin{table*}
\begin{center}
\begin{tabular}{l|c|cccccc} \hline
    Pre-training & Res & HMDB51 & UCF101 & DIVING48 & EGTEA & VOLLEY & YUP \\ \hline\hline

    Scratch & 112 & 31.5  & 70.3  & 4.7 & 50.5 & 53.8 & 43.7 \\ \hline
 
    Fractal & 112 & 65.4 & 87.6 & 40.3 & 56.0 & 83.1 & 72.7 \\ \hline\hline

    Scratch & 224 & 35.1 & 76.9 & 10.6 & 53.4 & 55.7  & 48.3 \\ \hline

    Fractal & 224 & 66.5 & 90.8 & \textbf{41.2} & 59.9 & \textbf{87.6} & 78.2\\ \hline
    Kinetics & 224 & \textbf{70.1} & \textbf{95.3} & 40.9 & \textbf{64.4} & 84.8 & \textbf{86.9} \\ \hline

\end{tabular}
\end{center}
\caption{Experiment 5 - Effectiveness of higher spatial resolution. Increasing the resolution boosts accuracy across all benchmarks. An interesting observation is that DIVING48 and VOLLEYBALL, where synthetic data surpasses Kinetics, exhibit the smallest variance of background amongst the evaluated datasets.}
\label{tab:exp5-resolution}
\end{table*}

As evidenced by Table \ref{tab:exp5-resolution}, increasing the resolution of fractals from $112$ to $224$ leads to nontrivial improvement across all benchmarks. This is reasonable as real videos often contain details such as small objects, which cannot be displayed adequately with lower resolution. Additionally, it can be observed that synthetic pre-training  surpasses Kinetics on the benchmarks DIVING48 and VOLLEYBALL. As seen in Fig. \ref{fig:downstream}, in the former, all videos display swimming pools and their surroundings such audience seats while the latter is exclusively set in volleyball courts. As such, the attribute that distinguishes the datasets where fractals surpass Kinetics is a small variance of the displayed background.

Nonetheless, synthetic data still lags behind Kinetics on the majority of evaluated benchmarks. 
However, it has to be reminded that the fractals were constructed automatically and therefore mitigate collection and annotation costs required for Kinetics. 
Moreover, it is notable that the fractal-pre-trained weights were obtained with less computation and time compared to what it would take for the Kinetics counterpart. To equalize the required training time and resources, the number of synthetic videos could be further increased.

\textbf{Experiment 6 - Something-Something V2}

To provide a more robust analysis and offer greater utility for our proposed approach, we further conduct experiments involving Something-Something V2 (SSv2) \cite{goyal2}. Compared to most previously evaluated datasets, SSv2 is significantly larger while simultaneously being more dependant on motion rather than appearance. In the added experiment, we compare pre-training with fractals to training from scratch as well as Imagenet weight  initialization, which is the standard approach for SSv2. Aside from TSM, results are additionally provided for the I3D architecture \cite{carreira1}.  I3D differs significantly from TSM, as the former is a 3D CNN while the latter is a 2D one. As such, we believe that these models should be sufficient to demonstrate the generalizability of our approach.

\begin{table*}
\begin{center}
\begin{tabular}{l|cccccc} \hline
    Pre-training & TSM & I3D \\ \hline\hline

    Scratch & 55.8 & 44.5 \\ \hline

    Imagenet & 58.8 & 51.2 \\ \hline

    Fractal & \textbf{59.7} & \textbf{52.6} \\ \hline

\end{tabular}
\end{center}
\caption{Experiment 6 - Impact of pre-training on SSv2. Fractal videos outperform all alternatives.}
\label{tab:exp6-ssv2}
\end{table*}

As shown in Table \ref{tab:exp6-ssv2}, fractals outperform Imagenet by $\sim$1\% for both models. Hence, it can be  deduced that, compared to Imagenet, our proposed synthetic data can produce more powerful 
neural representations for motion-related datasets such as SSv2. A possible explanation is that synthetic videos contain temporal patterns, while Imagenet is composed of static images. Furthermore, the accuracy improvement after pre-training is less perceptible compared to previous experiments. This can be  justified by SSv2's size, which is significantly larger than all the previous datasets. As the size of the downstream dataset increases, the models can develop more robust and generalized features without the need for pre-training. This tendency is similarly confirmed by the difference between HMDB51 \& UCF101 in previous experiments. While both contain videos of similar nature, the latter is twice as large but achieves half the accuracy increase from pre-training compared to the former ($\sim$15 \& 30\%).

\subsection{Manual Error Analysis}
\label{sec:manual_error_analysis}

Upon manual inspection of miss-classified samples from downstream datasets, a recurring characteristic can be distinguished. 
In particular, models pre-trained with fractals struggle with videos whose label is dictated  not by global information which covers the entire screen but by small details which occupy only a few pixels. A notable instance is small objects that are handled by humans. Examples are categories \say{Throw}, \say{Swing Baseball}, \say{Brush Teeth} and \say{Hammering} from datasets HMDB51 and UCF101. Likewise, the same holds for the entirety of EGTEA which displays cooking tools and ingredients. This justifies the ineffectiveness of synthetic pre-training:  only a 6\% accuracy gain is achieved compared to training from scratch. Another case involves facial expressions and motion of the mouth, as evidenced by categories \say{Smoke}, \say{Eat}, \say{Drink}, \say{Smile} and \say{Laugh} from HMDB51. Similarly, models are unsuccessful with limb movement. 
This is demonstrated again by EGTEA as well as the HMDB51 classes \say{Clap}, \say{Wave}, \say{Punch} and \say{Kick}. Frames from the aforementioned examples are displayed in Appendix Fig. \ref{fig:misclassified_frames}.

\begin{figure*}[t]
    \includegraphics[width=\linewidth]{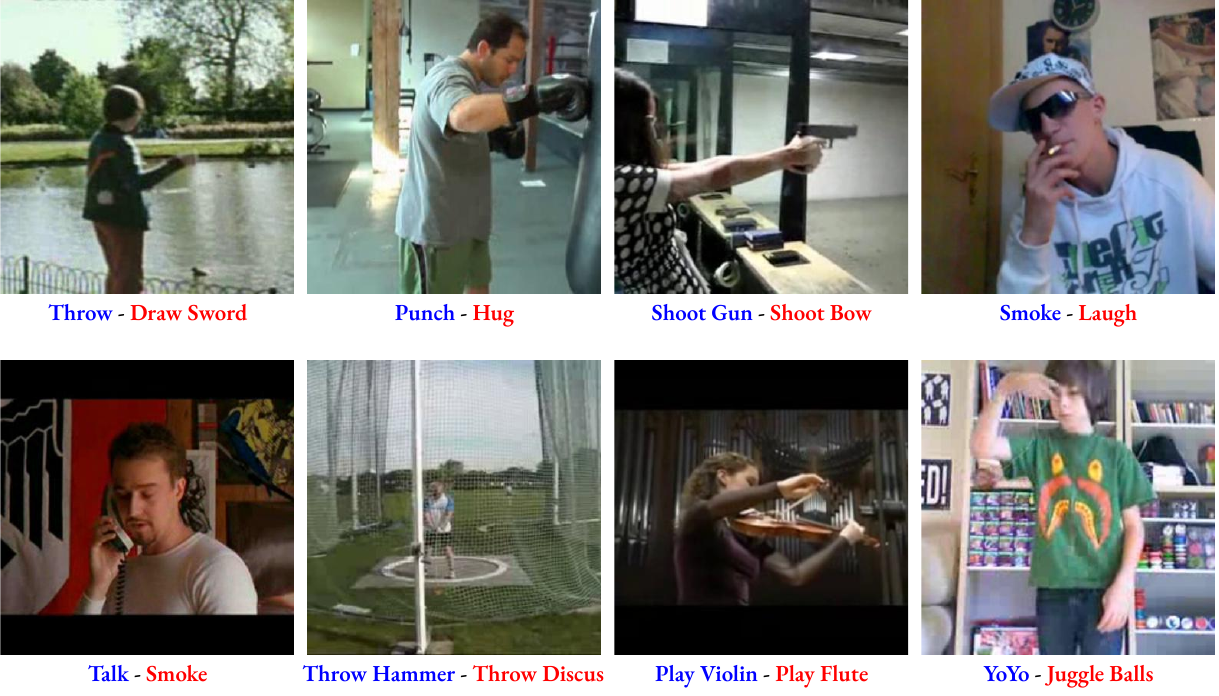}
    \caption{Frames from misclassified videos. \textcolor{blue}{Blue color} indicates ground truth, whereas \textcolor{red}{red color} indicates the model's incorrect prediction. In such videos the label is often determined by subtle details that cover a small percentage of the overall pixels. As such, the model fails to differentiate between similar categories.}
    \label{fig:misclassified_frames}
\end{figure*}

This deficiency in not unexpected. In our proposed synthetic datasets, the label, which is necessary for the training objective,  depends exclusively on the overall displayed fractal formation. Cases where the label is determined by local details are nonexistent. Consequently, the proposed pre-training does not adequately prepare CNNs for the aforementioned real-world instances. This shortfall  should be addressed by designing alternative generative processes in future work. Specifically, mirroring the conditions observed in real-world data, the constructed synthetic datasets should incorporate videos where the label is conditioned on a small percentage of pixels.

%% file: 4-related-work.tex
\section{Related Work}
\label{sec:related_work}

\textbf{Formula-driven synthetic data in computer vision.}
The work of \cite{kataoka1} employed fractal geometry to automatically generate large-scale labeled image datasets  to pre-train image models. Although their reported metrics are inferior to ImageNet pre-training \cite{russakovsky1}, they evidently surpass training from scratch. Subsequent work has proposed a more intuitive augmentation policy \cite{anderson1} and  demonstrated that the framework is compatible with different neural architectures \cite{nakashima1}. More recently, \cite{kataoka4} designed an alternative image synthesis method which results in representations that exceed ImageNet pre-training on specific model architectures. As an orthogonal approach, it has been demonstrated \cite{baradad1} that synthetic data results in strong representations only when certain conditions are met, such as replicating the attributes of real data. Thus, the design of synthetic data should be approached with meticulous care.

Synthetic data has also been utilized in other tasks. In particular, \cite{yamada1} pre-trained neural networks for point clouds with a synthetic datasets of 3D fractals. In the context of action recognition, \cite{kataoka3} pre-trained neural models with 3D perlin noise \cite{perlin1,perlin2}. Such videos are relevant to our work and thus serve as a baseline in later experiments. Additionally, \cite{zhao2} construct synthetic images of palm prints with the help of Bezier curves \cite{farin1}. 

Lastly, on a relevant note, synthetic data can alternatively be created with neural models such as Generative Adversarial Networks \cite{goodfellow2} and Diffusion Models \cite{ho2}. Indeed, it has been demonstrated that datasets of such synthetic images paired with unsupervised training objectives can produce impressive results that are on par or even superior to pre-training with real data \cite{fan1,tian1}. This methodology have also been shown to benefit greatly from scaling the synthetic data. However, compared to the formula driven counterpart, this approach is computationally more expensive and requires large amounts of real-world samples, suffering from the same limitations pertaining to real world datasets.

\textbf{Video Diffusion Models} 
Diffusion models \cite{ho2} are an emerging approach that has demonstrated impressive results in generation of images. More recently, diffusion has  been extended to the domain of video \cite{ho3}, exhibiting very promising results and great potential in many fields. This methodology bears similarity to the present work as it can also produce large datasets of synthetic videos.

However, despite the remarkable amount of new research, video diffusion models are not yet sufficiently mature \cite{ho3,melnik1}, posing multiple challenges that still need to be fully overcome. In the case of text-to-video models, a notable difficulty is the maintenance of temporal consistency. It has been observed that both the appearance and position of objects change wildly between video frames. This shortcoming has been constrained thanks to recent work, but not yet eliminated. Additionally, compared to our video synthesis methods, diffusion is typically significantly more demanding both in terms of resources and real-world training data. As such, video diffusion will not be investigated in the present work but remains an exciting prospect for future research.

\textbf{Action Recognition.} 
Contemporary action recognition models are first pre-trained on large-scale generic curated video datasets such as Kinetics \cite{carreira2,kay1},  Moments in Time \cite{monfort1} or YouTube 8M \cite{abu-el-haija1}. This process utilizes a supervised training objective. Alternatively, strong representations can also be obtained by utilizing the vast amount of available unlabelled videos and employing an unsupervised training objective \cite{feichtenhofer2}. Subsequently, transfer learning is employed and the models are fine-tuned on smaller specialized datasets such as UCF101 \cite{soomro1}, ActivityNet \cite{heilbron1} and HMDB51 \cite{kuehne1}. Omitting the first step significantly deteriorates downstream results. As such, this paper seeks to replace real large-scale video datasets with synthetic and automatically generated ones. 
Regarding neural architecture, the most common approaches are 2D \cite{simonyan1,wang5} and 3D \cite{carreira1,hara1,tran1} CNNs. However, recently, Vision Transformers \cite{tong1,wang6} have also begun to receive increased popularity.

\textbf{Applications of Fractals.} Owing to their aesthetic qualities, 2D fractals have been utilized in art \cite{draves1,draves2}. Furthermore, fractals are prominent in the field of image compression \cite{jacquin1}. It is noteworthy  that fractal geometry plays a significant role in domains outside of imaging including signal analysis \cite{maragos1,kokkinos1,dimakis1}, speech recognition \cite{maragos2,pitsikalis1,zlatintsi1} and telecommunications \cite{anguera1}. Lastly, due to their resemblance  to biological structures, fractals have been  employed in medical simulation \cite{costabal1,ionescu1}.

%% file: 5-conclusions.tex
\section{Discussion and Conclusion}

The present work automatically constructs synthetic datasets of short video clips. Such videos can be utilized for pre-training neural networks for the task of action recognition. Compared to real data, this approach eliminates the necessity for manual  dataset collection and annotation.  Additionally, in pursuit of minimizing the domain gap between real and synthetic videos we  introduce a set of heuristic domain adaptation techniques which mimic characteristics present in real data. The overall objective of our work is to determine properties of synthetic data as well as general guidelines that strengthen downstream performance. Observing experimental results, the following conclusions are reached:

\begin{itemize}
    \item Diversity of synthetic videos is a key factor for obtaining stronger visual representations and can boost  results regardless of the characteristics of the downstream dataset. This is in agreement with previous work \cite{baradad1}.

    \item Reducing the domain gap between real and synthetic videos also strengths downstream results. This can be achieved by emulating structural and motion-related properties of former during pre-training. Strict realism is not necessary and rough approximations of the said properties are sufficient. A few examples are background randomization, periodic motion and camera shaking.

    \item Supervised pre-training is a more cost-effective solution compared to self-supervised counterparts, achieving superior results under limited resources.

    \item Increasing the dataset size or spatial resolution consistently improves transferability.

\end{itemize}

It has to be noted that video action recognition is a field where access to video data on the scale of millions is currently relatively straightforward. In this regard, synthetic videos are unnecessary. However, we firmly believe that our findings are 
generalizable and can be transferred to other domains and tasks where available samples are scarce. A notable example could be medical video understanding, which could greatly benefit from synthetic data since obtaining real-world counterparts can be very expensive.
Nonetheless, the proposed methodology suffers from multiple limitations and improvements are expected to be added in future work. Specifically:

\begin{itemize}
    \item On the majority of evaluated benchmarks, synthetic pre-training lags behind Kinetics. We hypothesize that the gap in performance can be reduced, but not eliminated, by further increasing the quantity of synthetic training samples.

    \item Models pre-trained with synthetic data underperform in the detection of details, such as tools or facial expressions. As such, our proposed approach is not effective for datasets such as EGTEA GAZE+.  Future work should mitigate this  by constructing synthetic categories that are conditioned on a local cues and not global information, mirroring real data. 

    \item All synthetic videos produced in this work are of short length and depict a single motion. Thus, our framework is not suitable for applications involving longer videos, where the ability  to model contextual relation between distant frames is required.

\end{itemize}

%% file: ref.tex
\small
\bibliographystyle{spmpsci}
\bibliography{ref}

%% file: 6-appendix.tex
\clearpage

\section*{Appendices}

\appendix

\section{Kinetics Fine-tuning Hyperparameters}
\label{appendix:kinetics_hyps}

Table \ref{tab:appendix_kinetics_hyperparameters} lists the hyperparameters that were used to fine-tune the Kinetics checkpoint from \cite{lin1} in the final experiment.

\begin{table*}[t]
\begin{center}
\begin{tabular}{l|cccccc} \hline
    Hyperparameter & HMDB51 & UCF101 & DIVING48 & EGTEA & VOLLEY & YUP \\ \hline\hline

    Weight Decay  & $1 \cdot 10 ^ {-1} $ & $1 \cdot 10 ^ {-1} $ & $4 \cdot 10 ^ {-1} $ & $4 \cdot 10 ^ {-1} $ & $4 \cdot 10 ^ {-1} $ & $4 \cdot 10 ^ {-1} $ \\ \hline
 
    LR-Init & $1 \cdot 10 ^ {-8} $ & $1 \cdot 10 ^ {-8} $ & $2 \cdot 10 ^ {-7} $ & $2.5 \cdot 10 ^ {-8} $ & $5 \cdot 10 ^ {-8} $ & $5 \cdot 10 ^ {-8} $ \\ \hline
    
    LR-Peak & $1 \cdot 10 ^ {-5} $ & $1 \cdot 10 ^ {-5} $ & $2 \cdot 10 ^ {-4} $ & $2.5 \cdot 10 ^ {-5} $ & $5 \cdot 10 ^ {-5} $ & $5 \cdot 10 ^ {-5} $\\ \hline
    
    LR-Final  & $1 \cdot 10 ^ {-7} $ & $1 \cdot 10 ^ {-7} $ & $2 \cdot 10 ^ {-6} $ & $2.5 \cdot 10 ^ {-7} $ & $5 \cdot 10 ^ {-7} $ & $5 \cdot 10 ^ {-7} $ \\ \hline

\end{tabular}
\end{center}
\caption{Hyperparameters used for fine-tuning the Kinetics checkpoint.}
\label{tab:appendix_kinetics_hyperparameters}
\end{table*}

\section{Supplementary Visual Material}
\label{appendix:sup_visual_material}

\begin{figure*}[t]
    \includegraphics[width=\linewidth]{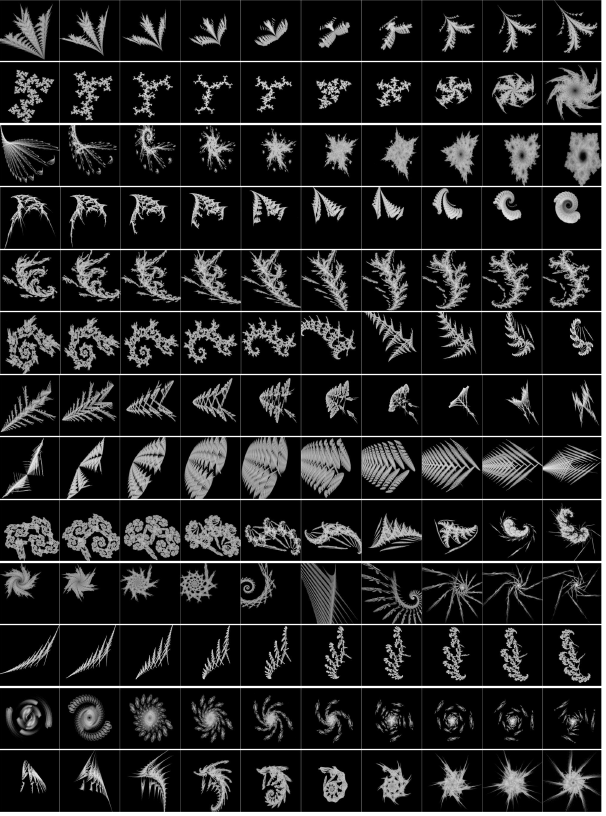}
    \caption{Examples of videos produced with Algorithm \ref{alg:decomposed_animation} using standard IFS. Each row displays frames from a different video.}
    \label{fig:appendix_fractal_video_linear}
\end{figure*}

\begin{figure*}[t]
    \includegraphics[width=\linewidth]{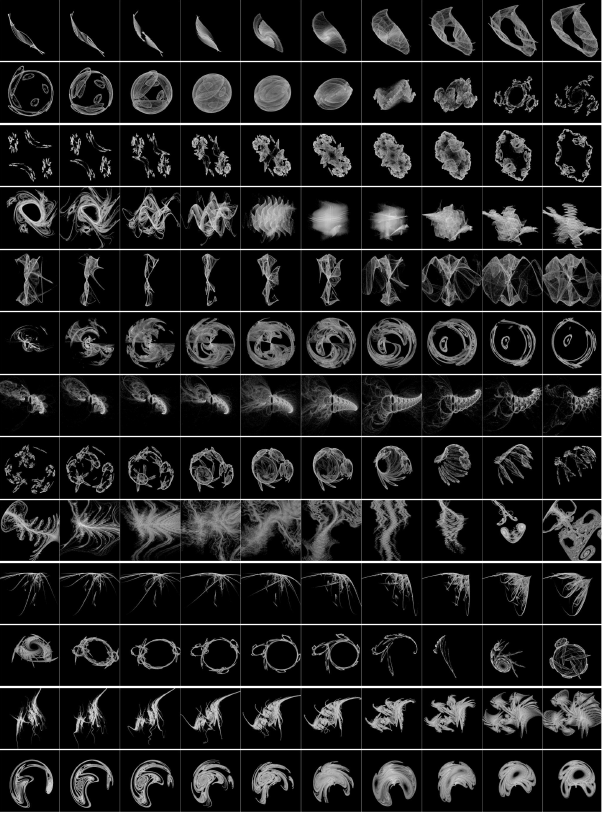}
    \caption{Examples of videos produced with Algorithm \ref{alg:decomposed_animation} using nonlinear IFS. Each row displays frames from a different video. Note the differences between Fig. \ref{fig:appendix_fractal_video_linear}.} \label{fig:appendix_fractal_video_nonlinear}
\end{figure*}

This section features additional visualizations of concepts described in the main document. Specifically:

\begin{itemize}

    \item Fig. \ref{fig:appendix_fractal_video_linear} contains examples of synthetic videos produced with Algorithm \ref{alg:decomposed_animation} using standard IFS.

    \item Fig. \ref{fig:appendix_fractal_video_nonlinear} contains examples of synthetic videos produced with Algorithm \ref{alg:decomposed_animation} using nonlinear IFS.

\end{itemize}

